\documentclass[10pt,twocolumn,letterpaper]{article}

\usepackage[pagenumbers]{wacv} 

\usepackage{graphicx}
\usepackage{amsmath}
\usepackage{amssymb}
\usepackage{booktabs}

\usepackage[utf8]{inputenc} 
\usepackage[T1]{fontenc}    
\usepackage{url}            
\usepackage{amsfonts}       
\usepackage{nicefrac}       
\usepackage{microtype}      
\usepackage{xcolor}         

\usepackage[pagebackref,breaklinks,colorlinks]{hyperref}

\usepackage[capitalize]{cleveref}
\crefname{section}{Sec.}{Secs.}
\Crefname{section}{Section}{Sections}
\Crefname{table}{Table}{Tables}
\crefname{table}{Tab.}{Tabs.}

\def\wacvPaperID{2618} 
\def\confName{WACV}
\def\confYear{2025}

\newcommand\tableimagewidth{0.108\textwidth}

\begin{document}

\title{Learning Complex Non-Rigid Image Edits from
Multimodal Conditioning}

\author{
  Nikolai Warner\\
  Georgia Institute of Technology\\
  Atlanta, GA \\
  {\tt\small nwarner30@gatech.edu}
  \and
  Jack Kolb\\
  Georgia Institute of Technology\\
  Atlanta, GA \\
  {\tt\small kolb@gatech.edu}
  \and
  Meera Hahn\\
  Google, Inc.\\
  Mountain View, CA \\
  {\tt\small meerahahn@google.com}
  \and
  Jonathan Huang\\
  Google, Inc.\\
  Mountain View, CA \\
  {\tt\small jonathanhuang@google.com}
  \and
  Irfan Essa\\
  Google, Inc. \&\\
  Georgia Institute of Technology\\
  Atlanta, GA \\
  {\tt\small irfan@gatech.edu}
  \and
  Vighnesh Birodkar\\
  Google, Inc.\\
  Mountain View, CA \\
  {\tt\small vighneshb@google.com}
}

\maketitle


\begin{abstract}
In this paper we focus on inserting a given human (specifically, a single image of a person) into a novel scene.  Our method, which builds on top of Stable Diffusion, yields natural looking images while being highly controllable with text and pose.  To accomplish this we need to train on pairs of images, the first a reference image with the person, the second a “target image” showing the same person (with a different pose and possibly in a different background).  Additionally we require a text caption describing the new pose relative to that in the reference image. In this paper we present a novel dataset following this criteria, which we create using pairs of frames from human-centric and action-rich videos and employing a multimodal LLM to automatically summarize the difference in human pose for the text captions. We demonstrate that identity preservation is a more challenging task in scenes “in-the-wild”, and especially scenes where there is an interaction between persons and objects. Combining the weak supervision from noisy captions, with robust 2D pose improves the quality of person-object interactions.
\end{abstract}

\section{Introduction}

\begin{figure}[t]
\centering
\begin{tabular}{@{\hskip 0.5pt}c@{\hskip 0.5pt} @{\hskip 0.5pt}c@{\hskip 10pt} @{\hskip 0.5pt}c@{\hskip 0.5pt} @{\hskip 0.5pt}c@{\hskip 0.5pt} @{\hskip 0.5pt}c@{\hskip 0.5pt}}
Scene & Ref & Edit 1 & Edit 2 & Edit 3 \\

\includegraphics[width=0.19\columnwidth]{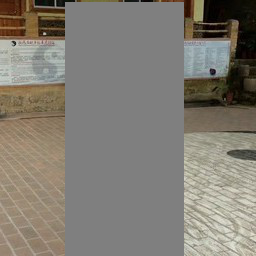} &
\includegraphics[width=0.19\columnwidth]{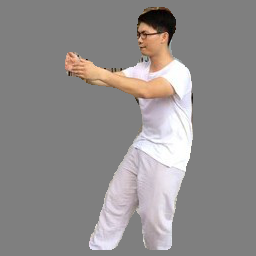} &
\includegraphics[width=0.19\columnwidth]{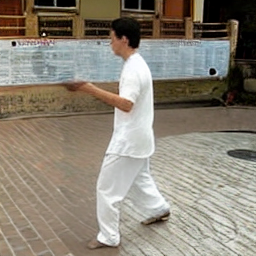} &
\includegraphics[width=0.19\columnwidth]{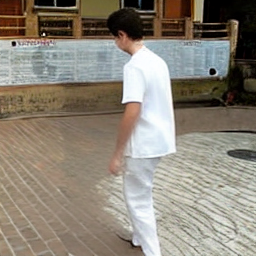} &
\includegraphics[width=0.19\columnwidth]{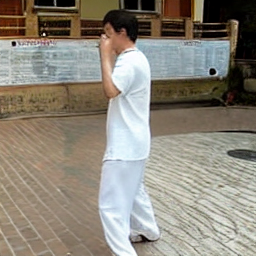} \\
\multicolumn{5}{p{0.95\columnwidth}}{\scriptsize \textit{Row 1: "He turns his body to the right and lowers his arms slightly." "He faces away showing his back side." "He raises his hand above his head."}} \\

\includegraphics[width=0.19\columnwidth]{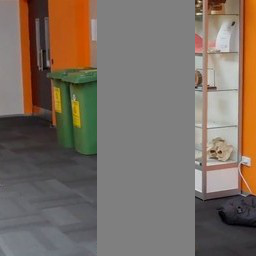} &
\includegraphics[width=0.19\columnwidth]{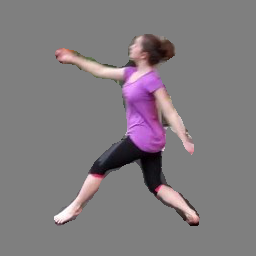} &
\includegraphics[width=0.19\columnwidth]{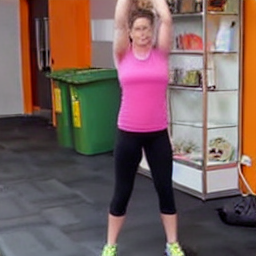} &
\includegraphics[width=0.19\columnwidth]{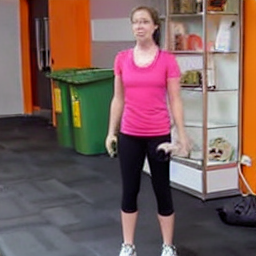} &
\includegraphics[width=0.19\columnwidth]{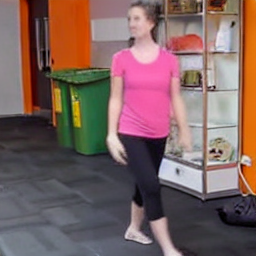} \\
\multicolumn{5}{p{0.95\columnwidth}}{\scriptsize \textit{Row 2: "She raises her arms above her head." "She lowers her arms." "She lunges her right leg forward."}} \\

\includegraphics[width=0.19\columnwidth]{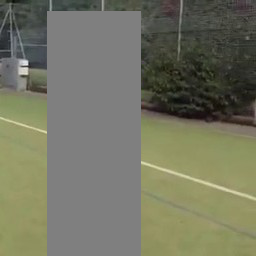} &
\includegraphics[width=0.19\columnwidth]{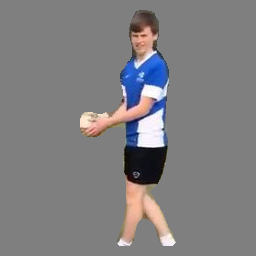} &
\includegraphics[width=0.19\columnwidth]{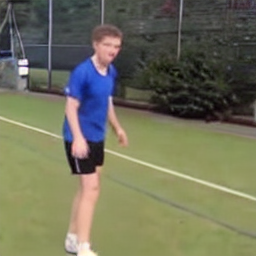} &
\includegraphics[width=0.19\columnwidth]{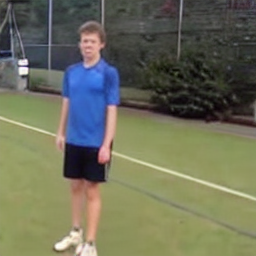} &
\includegraphics[width=0.19\columnwidth]{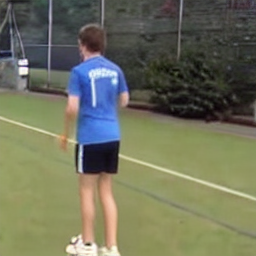} \\
\multicolumn{5}{p{0.95\columnwidth}}{\scriptsize \textit{Row 3: "He bends down reaching towards his toes." "He lowers his arms by his side." "He runs away with the ball."}} \\

\end{tabular}
\caption{Given a single image, multiple controllable identity-preserving edits can be specified with different text captions. Given a masked insertion scene and reference image containing a person to insert, our fine-tuned model inserts them into the scene controllable as controlled by a given text caption. Text and image-based inference on unseen images using an image + text model. Where captions are unavailable, we train the text and image model on a blank caption.}
\label{fig:multiple_edits_same_photo}
\end{figure}

\begin{figure}[t!]
\centering
\begin{tabular}{@{\hskip 0.5pt}c@{\hskip 0.5pt}@{\hskip 0.5pt}c@{\hskip 0.5pt}@{\hskip 0.5pt}c@{\hskip 0.5pt}@{\hskip 0.5pt}c@{\hskip 0.5pt}}
Scene & Ref & Ours & Ground Truth \\

\includegraphics[width=0.23\columnwidth]{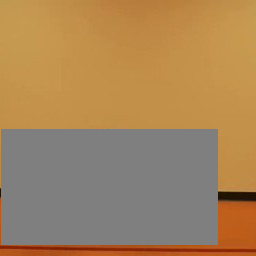} &
\includegraphics[width=0.23\columnwidth]{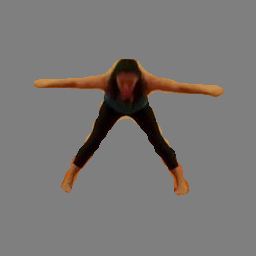} &
\includegraphics[width=0.23\columnwidth]{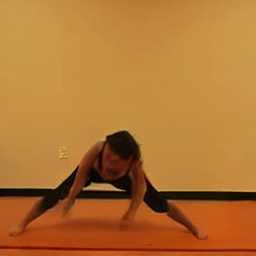} &
\includegraphics[width=0.23\columnwidth]{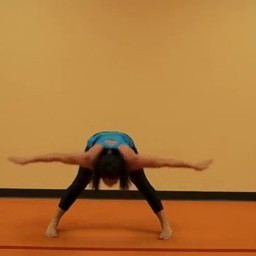} \\
\multicolumn{4}{p{0.95\columnwidth}}{\scriptsize\textit{Row 1: "She bends forward at the waist, touching the ground with her hands spread apart."}} \\

\includegraphics[width=0.23\columnwidth]{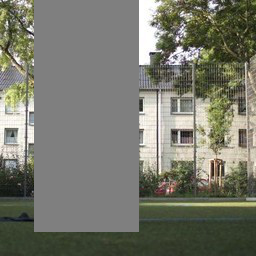} &
\includegraphics[width=0.23\columnwidth]{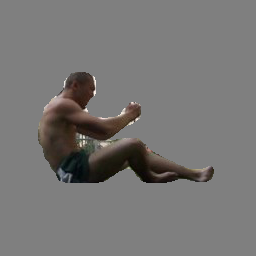} &
\includegraphics[width=0.23\columnwidth]{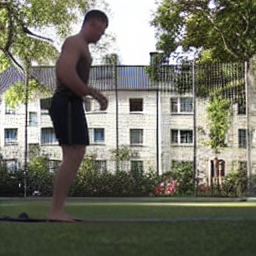} &
\includegraphics[width=0.23\columnwidth]{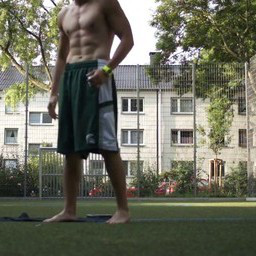} \\
\multicolumn{4}{p{0.95\columnwidth}}{\scriptsize\textit{Row 2: "He goes from sitting on the ground to standing up."}} \\

\includegraphics[width=0.23\columnwidth]{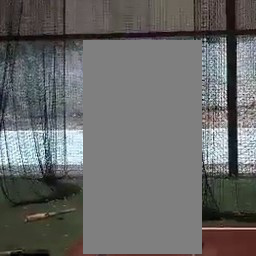} &
\includegraphics[width=0.23\columnwidth]{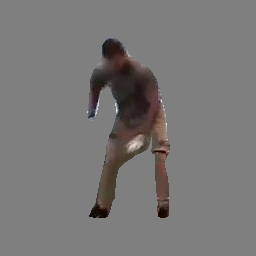} &
\includegraphics[width=0.23\columnwidth]{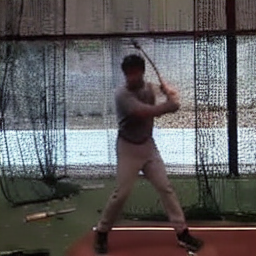} &
\includegraphics[width=0.23\columnwidth]{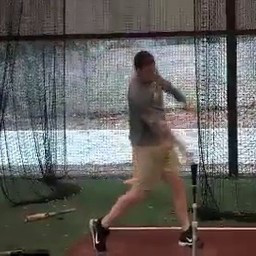} \\
\multicolumn{4}{p{0.95\columnwidth}}{\scriptsize\textit{Row 3: "He follows through after swinging the baseball bat, moving it to his left side."}} \\

\includegraphics[width=0.23\columnwidth]{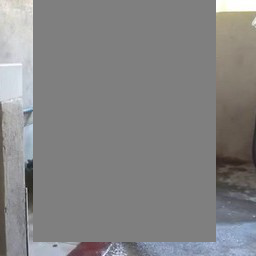} &
\includegraphics[width=0.23\columnwidth]{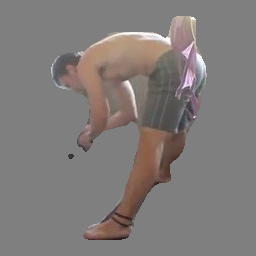} &
\includegraphics[width=0.23\columnwidth]{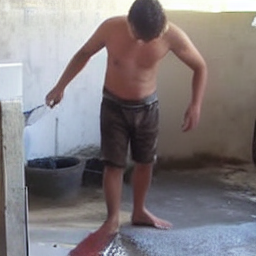} &
\includegraphics[width=0.23\columnwidth]{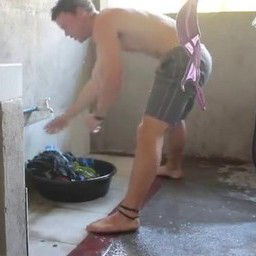} \\
\multicolumn{4}{p{0.95\columnwidth}}{\scriptsize\textit{Row 4: "He stands upright, looking down at the clothes in the basin."}} \\
\end{tabular}
\caption{Complex edits are achievable through weakly annotated supervision of a portion of the overall dataset, and finetuning jointly on text. Given a masked scene to insert a person into ("Scene") and a segmented crop of the person ("Ref"), the person is inserted into the scene. For comparison, we also provide the ground truth image that is paired with the caption and input. Given a scene, and person to insert, our model is capable of multiple non-rigid edits that preserve the identity of the person, despite a relatively small set of 13,487 weakly annotated captioned image pairs out of 78,000 videos. See Appendix Table 1 for dataset details and Section \ref{sec:three_three} for details on our weakly annotated captions.}
\label{fig:complex_nonrigid}
\end{figure}

Generative diffusion models are an essential tool for creating high-quality images, videos, and 3D models. For applications involving human subject matter, the models can edit images in a controlled manner through textual and pose inputs. This capability allows for precise, user-defined modifications without the need for detailed manual editing. In this paper, we explore the use of generative diffusion models for conducting complex, non-rigid image edits that represent a challenging class of modifications.

Despite their capabilities, current generative models do not fully leverage the extensive implicit world knowledge they accrue during training. By directing these models to manipulate this knowledge through structured edits explicitly, we aim to expand their practical applications in image editing. Our results show improvements in the model's ability to preserve identity and contextual accuracy in generated images across controlled and uncontrolled environments. This work advances the capabilities of generative image editing, and lays groundwork for further developments in intuitive, user-centric image editing technologies.

Prior work has organized the complexity of human subject image editing into rigid and non-rigid edits \cite{kulal2023putting}. Rigid edits preserve the subject's pose and alter their appearance (e.g., changing their shirt), while non-rigid edits modify the pose and may preserve appearance (e.g., generating yoga poses from a reference). Non-rigid edits are complex as they involve deforming the subject in a natural and coherent manner. Increasing the difficulty, non-rigid edits can include object interactions, requiring the model to know how to interact with the object's class. For user applications, open-vocabulary text guidance is desirable for prompting edits.

Central to our approach is the integration of multi-modal large language models (LLMs) with video data, enabling a novel form of noisy, self-supervised learning. This integration allows our models to learn text-controllable edits from less structured video data, approximating the unpredictability of real-world scenarios. Using video clips as datasets enriches the model's understanding of human motion and interactions across various contexts, enhancing its capability to manage complex edits and improving its comprehension of human-object interactions within scenes.

This paper contributes the following:
\begin{enumerate}
\item \textbf{Controllable Non-Rigid Edits on Real World:}  We enable non-rigid edits of human subject scenes, controllable by text, on in-the-wild data and public datasets. We show considerable qualitative improvements over baseline models and algorithms.

\item \textbf{Training with Language-Guided Controllability:} We develop a new conditioning pipeline that uses multimodal LLMs to provide noisy guidance through world knowledge to condition and model non-rigid insertions.

\item \textbf{Superior Person-Object Interactions}: We improve controlling interactions between objects and persons. Our model integrates combinations of reference images, textual descriptions, and pose data, showing superior performance compared to baselines that are image-only, image-text, or image-pose. 

\item \textbf{Open-Source Annotated Dataset}: We collect a human-annotated dataset of thousands of captions and processed video frames from multiple person-centric sources. We also validate prior work and re-train baselines that were originally trained on closed-source data. 
\end{enumerate}

\section{Related Works}

\begin{figure*}[t!]
\centering
\begin{tabular}{@{\hskip 0pt}c@{\hskip 10pt}c@{\hskip 1pt}c@{\hskip 1pt}c@{\hskip 10pt}c@{\hskip 1pt}c@{\hskip 10pt}c@{\hskip 1pt}c@{\hskip 0pt}}
\multicolumn{1}{c}{} & \multicolumn{3}{c}{Non-Diffusion Approaches} & \multicolumn{1}{c}{} & \multicolumn{3}{c}{Affordance-Diffusion Approaches} \\
Input Image & MASACtrl & ControlNet & PIDM* & Masked Input & Reference Crop & \textbf{Our's} & Kulal et al.† \\

\includegraphics[width=\tableimagewidth]{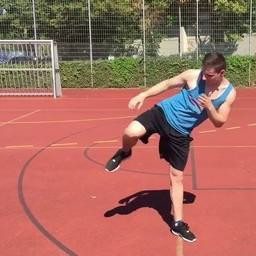} &
\includegraphics[width=\tableimagewidth]{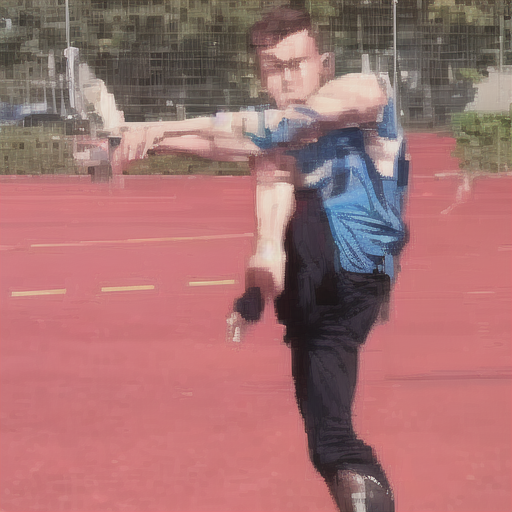} &
\includegraphics[width=\tableimagewidth]{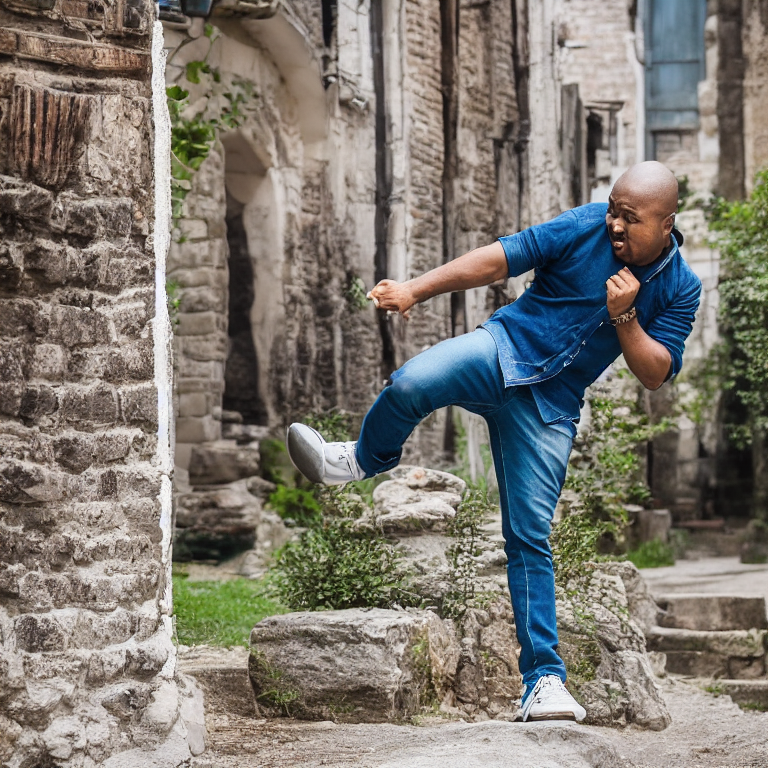} &
\includegraphics[width=\tableimagewidth]{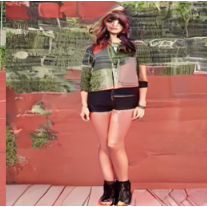} &
\includegraphics[width=\tableimagewidth]{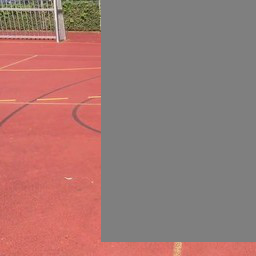} &
\includegraphics[width=\tableimagewidth]{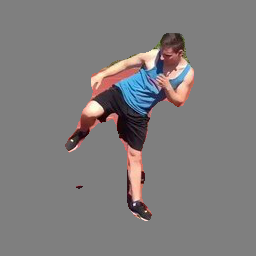} &
\includegraphics[width=\tableimagewidth]{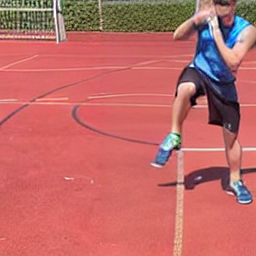} &
\includegraphics[width=\tableimagewidth]{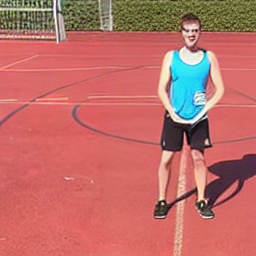} \\
\multicolumn{8}{c}{↑ control prompt: ``\textit{He kicks his right leg up higher while leaning back.}"} \\

\includegraphics[width=\tableimagewidth]{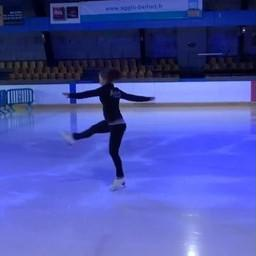} &
\includegraphics[width=\tableimagewidth]{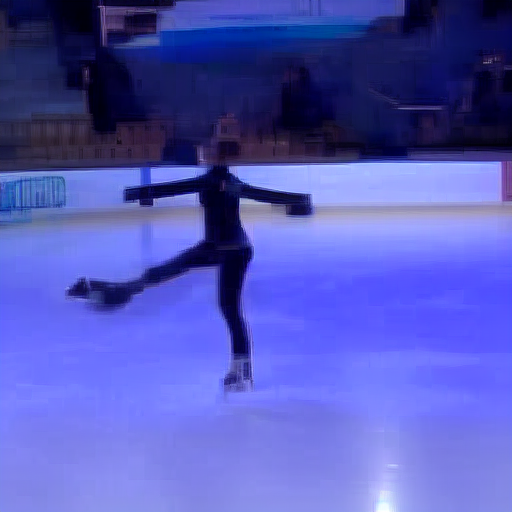} &
\includegraphics[width=\tableimagewidth]{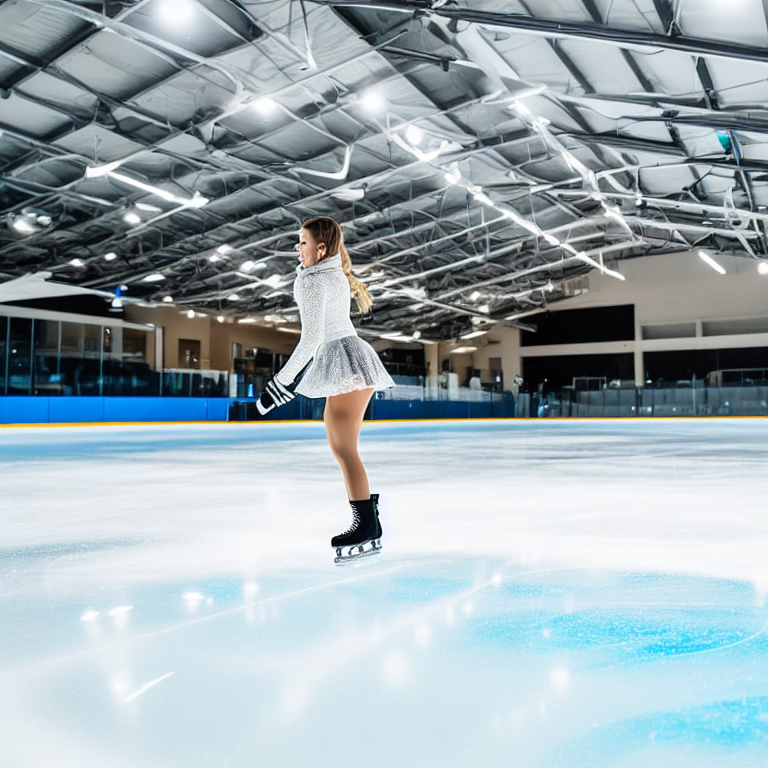} &
\includegraphics[width=\tableimagewidth]{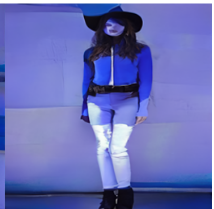} &
\includegraphics[width=\tableimagewidth]{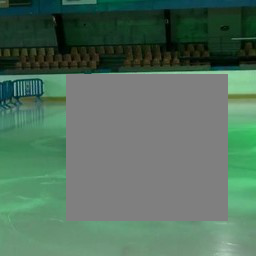} &
\includegraphics[width=\tableimagewidth]{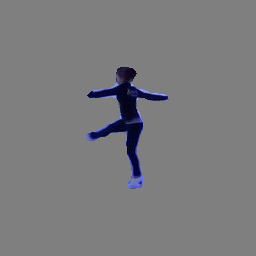} &
\includegraphics[width=\tableimagewidth]{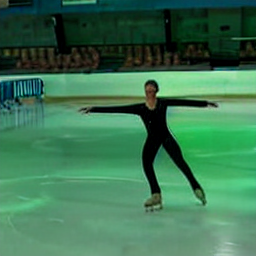} &
\includegraphics[width=\tableimagewidth]{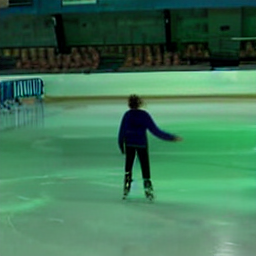} \\
\multicolumn{8}{c}{↑ control prompt: ``\textit{She extends her right leg back and arms outwards while ice skating.}"} \\

\includegraphics[width=\tableimagewidth]{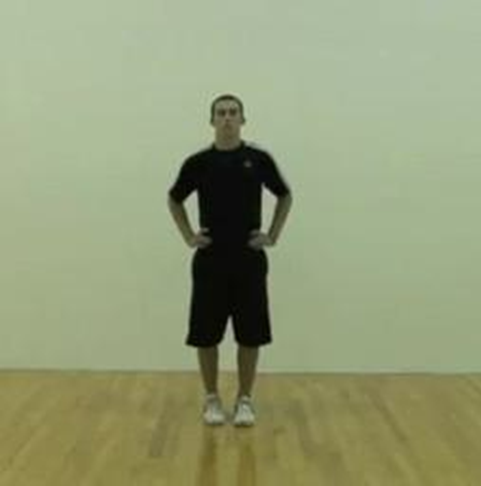} &
\includegraphics[width=\tableimagewidth]{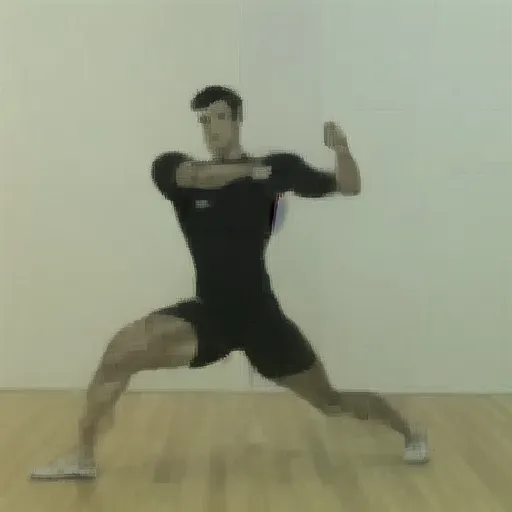} &
\includegraphics[width=\tableimagewidth]{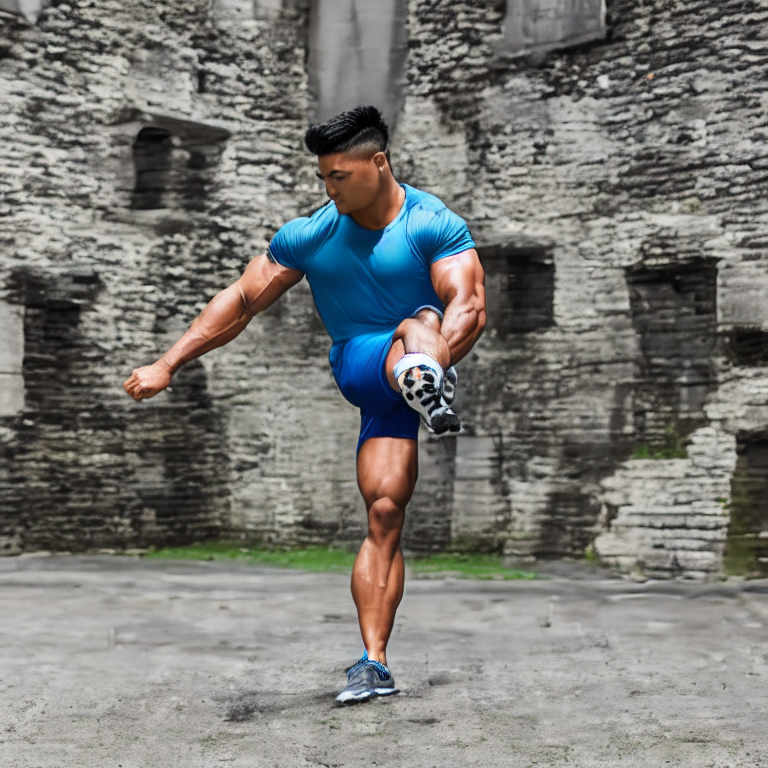} &
\includegraphics[width=\tableimagewidth]{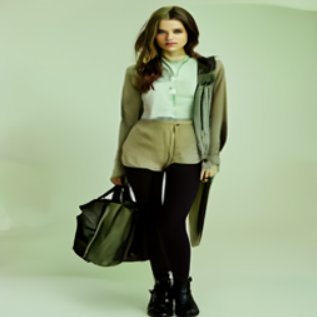} &
\includegraphics[width=\tableimagewidth]{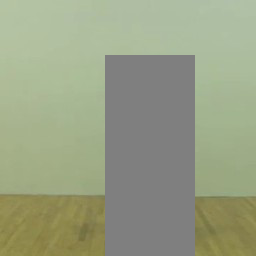} &
\includegraphics[width=\tableimagewidth]{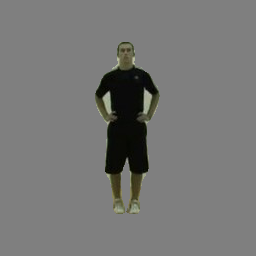} &
\includegraphics[width=\tableimagewidth]{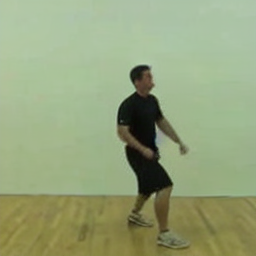} &
\includegraphics[width=\tableimagewidth]{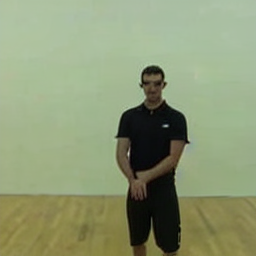} \\
\multicolumn{8}{c}{↑ control prompt: ``\textit{He lunges with his right leg.}"} \\

\includegraphics[width=\tableimagewidth]{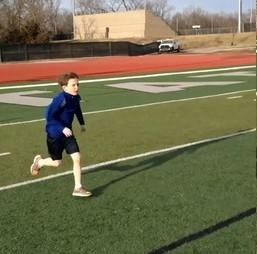} &
\includegraphics[width=\tableimagewidth]{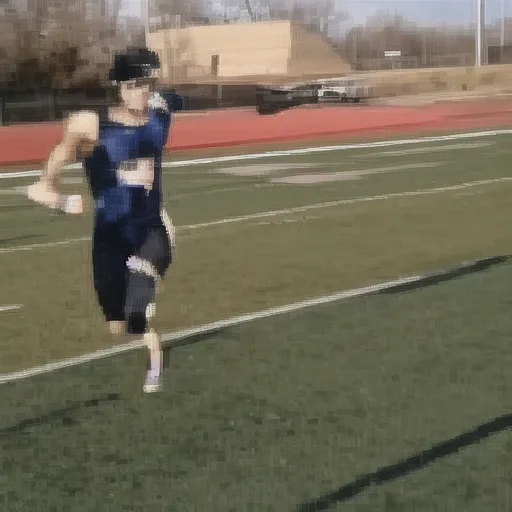} &
\includegraphics[width=\tableimagewidth]{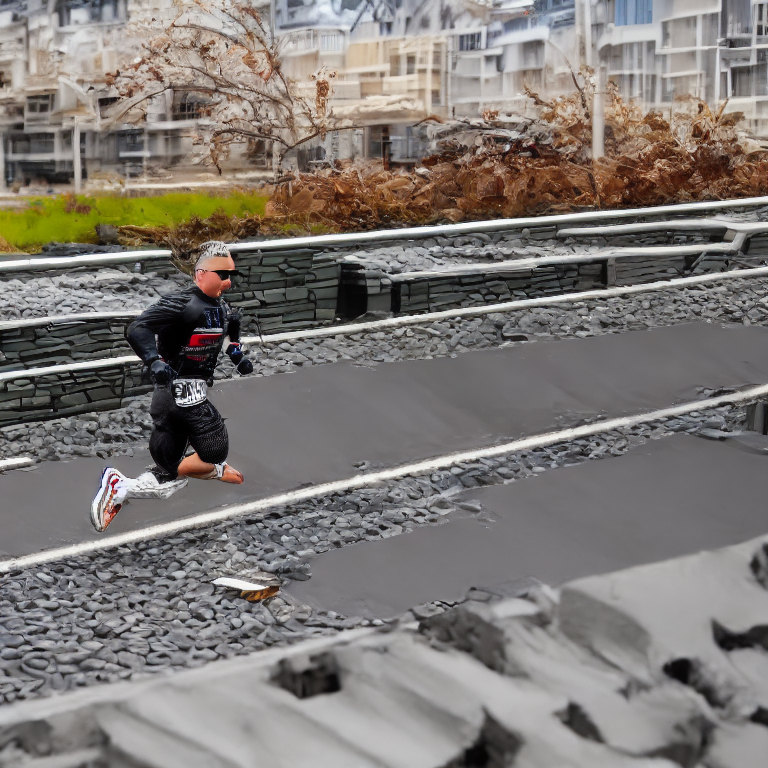} &
\includegraphics[width=\tableimagewidth]{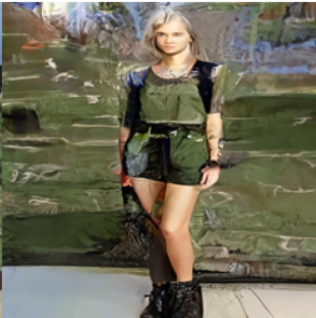} &
\includegraphics[width=\tableimagewidth]{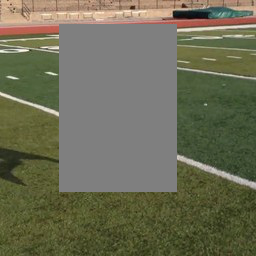} &
\includegraphics[width=\tableimagewidth]{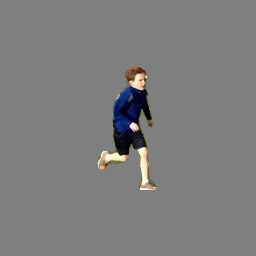} &
\includegraphics[width=\tableimagewidth]{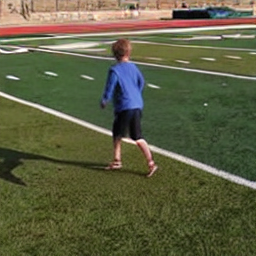} &
\includegraphics[width=\tableimagewidth]{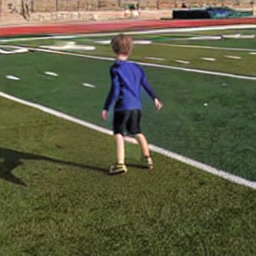} \\
\multicolumn{8}{c}{↑ control prompt: ``\textit{He runs towards the left.}"} \\

\includegraphics[width=\tableimagewidth]{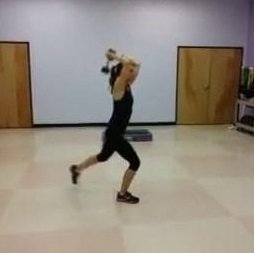} &
\includegraphics[width=\tableimagewidth]{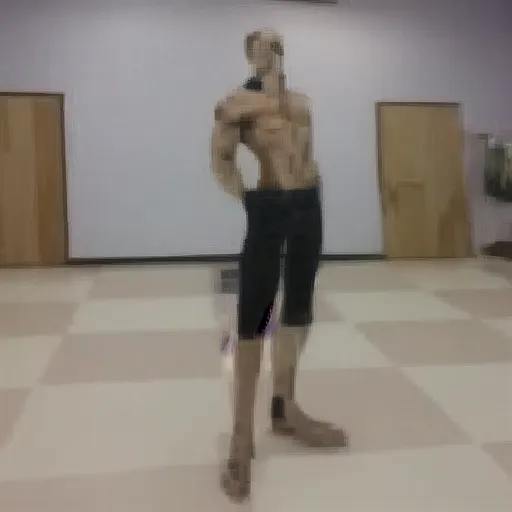} &
\includegraphics[width=\tableimagewidth]{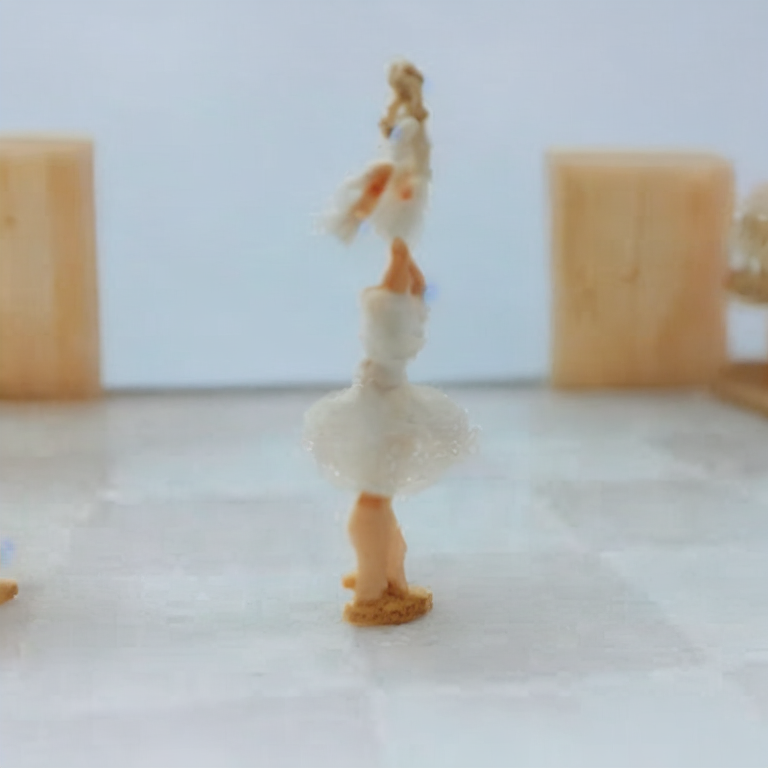} &
\includegraphics[width=\tableimagewidth]{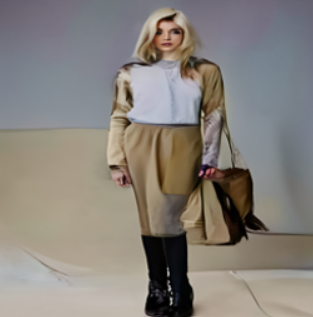} &
\includegraphics[width=\tableimagewidth]{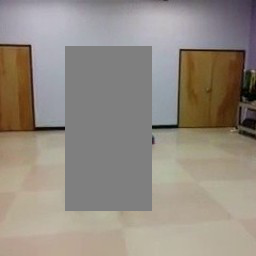} &
\includegraphics[width=\tableimagewidth]{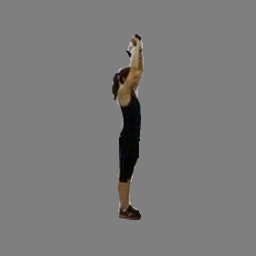} &
\includegraphics[width=\tableimagewidth]{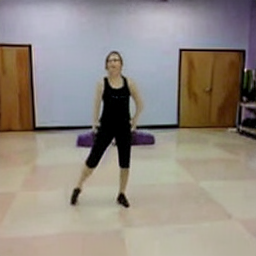} &
\includegraphics[width=\tableimagewidth]{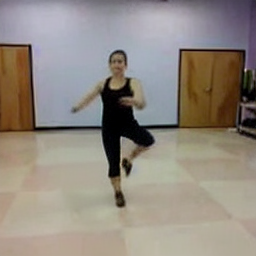} \\
\multicolumn{8}{c}{↑ control prompt: ``\textit{She holds her hands on her hips.}"} \\
\end{tabular}
\caption{Comparison of our approach to baselines for identity preservation and controllability of in-the-wild images. The input image is controlled using the prompts below each row, and the human subject is transferred to new frames for relevant models (our's and Kulal et al.). No baseline achieves comparable identity-preserving non-rigid edits on in-the-wild data. Baselines either insert a person without controllability (Kulal et al.), or are controllable but fail to generalize to in-the-wild images (MASACtrl, PIDM). Our approach maintains similar photorealism to Kulal et al., with improved controllability.\\
* PIDM works well on its training dataset but is brittle in the wild, likely due to its fashion-related training dataset.\\
† Kulal et al. results are from the re-trained model with image conditioning only.}
\label{fig:nonrigid_comparison}
\end{figure*}

Recent years have seen high research interest in improving the visual quality and controllability of images produced from generative models.

\subsection{Text-Based Control \& General Controllability}

The original latent diffusion paper introduced conditional control through the use of cross-attention layers in the U-Net \cite{ronneberger2015unet}. Combined with performing diffusion in the latent space with smaller spatial dimensions, this enabled a variety of control signals and auxiliary tasks including text-to-image, layout-to-image, inpainting, and super-resolution. 

Prior works have approached text-based control of images. DreamBooth allows editing various image elements, such as the background, style, and accessories, while maintaining the image's primary subject \cite{zhang2022dreambooth}. The model's authors fine-tuned existing text-to-image diffusion models using a few-shot dataset of the subject to insert them into new domains. Imagic introduced a three-stage approach to enable non-rigid pose edits by performing linear interpolation between the optimized and target text embeddings \cite{kawar2022imagic}.

ControlNet aims to address the issue of catastrophic forgetting \cite{zhang2023controlnet}. The model uses a reference image and a text prompt, and is capable of non-rigid edits, however it does not preserve identity. MASACtrl enables text-driven nonrigid edits in a tuning-free manner via “masked mutual attention”, however is sensitive to parameters chosen to control where in the diffusion process the masked mutual attention occurs \cite{cao2023masactrl}. We use ControlNet and MASACtrl as baselines in this work.

\subsection{Pose-Based Control}
Few works have approached non-rigid edits given a target pose. PIDM incorporates a noise prediction model and a texture encoder to maintain the subject's style given a target pose \cite{bhunia2023person}. The model is trained on DeepFashion \cite{liu2016deepfashion} with \(\sim\)52,000 images, however is brittle outside that domain. PCDM fine-tunes a pre-trained latent diffusion model \cite{shen2023advancing}. The model aligns the source image and target pose through a three-layer trainable pose network to project the source and target poses into a latent pose embedding. We use PIDM as a baseline for this work.

\subsection{Identity-Preserving Diffusion}
Recent advancements in identity-preserving diffusion models have maintained subject identities while enabling detailed and flexible image editing. InstantID uses a novel IdentityNet module that integrates spatial conditioning with a diffusion model for identity preservation using a single facial image \cite{wang2024instantid}. He et al. introduced regularization in the dataset generation, enhancing identity preservation across various text-to-image models \cite{he2024data}. Banerjee et al. proposed a technique for simulating aging and de-aging in face images using latent text-to-image diffusion models, maintaining biometric identity with high photorealism \cite{banerjee2023identity}. These methods maintain subject identities, however are non-rigid edits.

\subsection{Non-Rigid Identity Preservation}
Only one prior work has accomplished non-rigid pose changes with identity preservation. Kulal et al. showed that an inpainting diffusion framework can train a model to hallucinate a plausible and photorealistic insertion of a person from one scene into another \cite{kulal2023putting}. For reference images, the model employs a nearest-neighbor approach where frames are sampled from a large meta-dataset of person and activity-centric videos \cite{brooks2022hallucinating}. Unfortunately, the model was trained on proprietary data and no model checkpoints are available.

To transform the algorithm into a viable baseline, we re-implement Kulal's methods on publicly available data. To enable controllability we train the model on automated captions and extracted poses from frames, and use additional filtering criteria to improve the quality of resulting images.

\section{Methods}

\begin{figure*}[t]
\centering
\includegraphics[width=\textwidth]{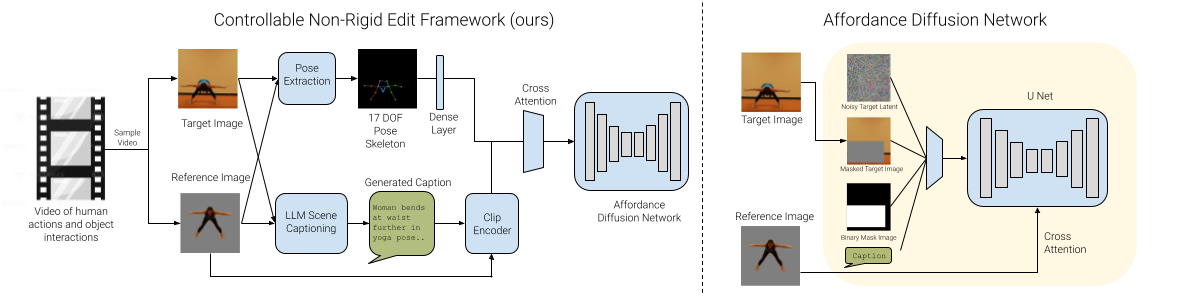}
\caption{System diagram illustrating the process of generating a desired edit using multiple inputs including noise target latent, binary mask, masked target latent, reference image, and change of scene prompt. The Affordance Diffusion Network on the right is the formulation proposed by \cite{kulal2023putting}, our improvements to controllability come from the framework described on the left. We study combinations of pose and weakly annotated text conditioning to learn more controllable and complex image edits that still preserve the identity of a person in a scene. No other work to our knowledge combines controllable non-rigid edits with identity preservation, and works in the wild. }
\label{fig:system_diagram}
\end{figure*}

\subsection{Preliminaries}

\subsubsection{Diffusion Models}
Diffusion models are a class of generative models that iteratively refine noisy samples to generate realistic images. The model starts with a random noise image and progressively denoises it, following a defined process, to produce a coherent output. In each step the model predicts and reduces noise from the previous step using a pre-trained noise estimation network.

The training objective is to minimize the difference between the predicted noise and the actual noise added during the forward process:
\[
L = \mathbb{E}_{t, \mathbf{x}_0, \mathbf{\epsilon}} \left[ \|\mathbf{\epsilon} - \mathbf{\epsilon}_\theta(\mathbf{x}_t, t)\|^2 \right]
\]

\subsubsection{Conditional Diffusion}
Conditional diffusion models enhance the basic diffusion process by incorporating additional information to guide the generation. This conditioning information can be in the form of text descriptions, class labels, or other metadata. The goal is to steer the diffusion process to generate images that meet specific criteria or exhibit particular characteristics defined by the conditioning input. Affordance diffusion models, a subset of conditional diffusion, generate images that reflect plausible interactions between objects and humans within a scene, learning from videos where human actions are contextually grounded.

\subsubsection{Latent Diffusion}
Latent diffusion models operate in a compressed, lower-dimensional latent space rather than the pixel space, making the generation process more efficient. By performing diffusion in this latent space, these models can handle higher resolution images and complex transformations with reduced computational cost. This approach leverages pre-trained autoencoders to map images to and from the latent space, maintaining high fidelity while optimizing processing time.




\subsection{Inpainting Diffusion Formulation}
We frame affordance diffusion learning as an inpainting diffusion finetuning problem. Person-centric video datasets are collected and processed such that only one person is present in each video and redundant frames with little motion are removed. We finetune an inpainting diffusion checkpoint from Stable Diffusion on sampled pairs of images from these videos.

Each video meets our pose detection criteria: a single pose skeleton must be present in each sampled scene, and the majority of joints must be visible. For each pair of sampled frames, one frame is used as the masked target where the person is masked out, and the other frame is cropped to include only the person. This cropped frame is then used to preserve the identity of the person via cross-attention mechanisms. By focusing on pairs of frames with significant motion, the model learns to inpaint the masked target image while maintaining the identity and pose of the person from the reference frame. For more details, refer to \cite{kulal2023putting}.

\subsection{Enabling Controllability from Noisy Supervision with Scene Difference Captions}
\label{sec:three_three}
Previous literature has trained models to hallucinate a plausible way to insert a person into a scene \cite{kulal2023putting, brooks2022hallucinating}. Given a scene with a masked area to inpaint, and a segmented person to insert, the model learns plausible ways to insert a person into a scene. However, this suffers from limited controllability. For example, there are multiple ways a child could interact with a slide in a playground.

We finetune a pre-trained Stable Diffusion inpainting checkpoint to make image edits that respect textual prompts. From preprocessing the Kinetics-700 dataset \cite{carreira2019kinetics} for pose detection and fidelity, we obtain 5,787 captioned videos. We also collect 7,700 annotated image pairs from videos in NTU-RGBD \cite{Shahroudy_2016_NTU_RGBD}, and caption pairs from Charades \cite{Zhang_2019_TRG} and Fit3D \cite{Fieraru_2021_CVPR}. We use GPT-4V \cite{bubeck2023sparks} in a 10-shot manner to caption the difference between scenes given two images. This is often an ill-defined problem, as depending on the degree of difference between two frames, conflicting captions are possible. For example, two frames could show dropping an object or reaching to pick it up. Images were sent as side-by-side composite images instead of successive images, as we found it reduced GPT-4V's hallucination.

\subsection{Enabling Performative Models on Limited Data}
A key challenge was training on limited publicly available data to create photorealistic edits with identity-preservation and faithfulness to conditioning signals.

To facilitate learning complex non-rigid edits that generalize to in-the-wild data, we sample a handful of key frames per video instead of retaining all frames. Each keyframe is sampled according to several criteria. First, we use pose detection to remove frames that contain more than one person, or do not contain most of a pose skeleton present. Many frames from Kinetics-700 contain close up videos of partial pose skeletons, such as a person from the neck up or a zoomed in video of a hand. In addition, many videos contained multiple perspectives, where initially a single person was present but the zoom level changed.

We also specify a minimum pose distance between sampled frames equivalent to the length of the pose skeletons shoulder to head distance. We found that using a single-stage model such as OpenPose underperformed a two-stage \textit{object detection + pose extraction model} such as RTMPose \cite{jiang2023rtmpose}. We use the pose information extracted during preprocessing to help guide the noisy captions during training for better person-object interactions. Image histogram similarity is used to set a minimum and maximum similarity between frames. Due to the small training dataset, frames must be similar enough that the background or context does not entirely change, but distinct enough to avoid redundant frames.

\subsection{Improving Person-Object Interactions}

Our objective is to learn complex non-rigid edits to in-the-wild person-centric scenes. Often, scenes involve human-object interactions, which has been unexplored in prior work.

Many GPT-4V captions reasonably describe a scene, or guess a likely action (“hitting a baseball bat”) but not the specific part of the action. The emphasis on broader context helps our model learn what parts of the image and person are important given a reference person and scene to insert. See the Appendix for examples.

To assist the model in learning a distribution of possible poses given an object, we add joint conditioning with pose data, as shown in Figure \ref{fig:system_diagram}. At inference time, the model receives a masked scene to insert a person into, a reference person from the same video, a bounding box describing where to insert the person, and a text prompt describing how the person's posture or object interaction changes. This is a similar process to InstructPix2Pix \cite{brooks2023instructpix2pix}, except the scene difference captions contain information about non-rigid deformations that do not respect the structure of the original person or photo.

We find that when we re-implement image-only conditioning as a baseline method, shown in Tables 2 and 3, identity preservation decreases when we focus on scenes that contain person-object interactions. This is because the model must focus on preserving the identity of the object. Because of our automated masking and segmentation pipelines, some object interactions may be described in the masked target image (e.g., half a bicycle), and some may only be described in the reference person crop (e.g., a kid holding a ball). When text prompts are generated they are relative to the unmasked target and reference images. Therefore, text and pose information are critical for successful modeling of person-object interactions.

\subsection{Inference Implementation Details}

\begin{table}[t!]
\centering
\begin{tabular}{@{}l|c|c|c|c@{}}
\hline
Condition & \multicolumn{2}{c}{FID} & \multicolumn{2}{c}{PCKh} \\ \hline
          & Train & Val & Train & Val \\ \hline
img (Kulal)*      & \textbf{38.70} & \textbf{45.07} & \textbf{0.196} & 0.0492 \\
img-pose  & 40.01 & 45.97 & 0.177 & 0.0619 \\
img-text  & 40.55 & 46.39 & 0.176 & 0.0471 \\
img-pose-text & 40.49 & 45.69 & 0.185 & \textbf{0.0625} \\ \hline
\end{tabular}
\vspace{0.05in}
\caption{Ablation of Img, Pose, Text; FID, PCKh on seen and unseen images. Img, Text, Pose performs the best on PCKh controllability, while the image only model achieves the best FID. However, FID does not measure identity preservation, hence see the discussion in Experiments. Thus, we measure identity preservation, and respect of control signals in Tables \ref{tab:user_study_non_object} and \ref{tab:user_study_object}. We trade a small amount of photorealism for greater controllability. We reimpliment Kulal et al. on publicly available data, as denoted by "*".}
\label{tab:FID,PCKh}
\end{table}

At inference time, we modify Classifier-Free Guidance (CFG) \cite{ho2022classifier} for each of our conditioning setups. For image-only conditioning, representing a re-implementation of the Kulal et al. baseline on our publicly available training data; we encode the unconditional representation as a tensor of all zeros in the same shape as the reference image conditioning with CLIP. The mask is expanded to cover the entire target image.

For image-text conditioning, we represent the unconditional reference image conditioning signal as a zeros tensor with the same shape as the reference image, encoded with CLIP. Then, we encode the unconditional representation of the scene difference caption as a null caption, encoded with CLIP.

For models with pose conditioning (image-pose, image-pose-text), we define an unconditional pose representation as follows: an image with a neutral posture from the dataset is selected (a person standing straight in the center of the frame, with their hands at their side) and passed through a linear projection layer to obtain the same embedding dimension as CLIP. The resulting unconditional pose representation is concatenated with the other unconditional signals.

\section{Experiments}

Our architecture is shown in Fig. \ref{fig:system_diagram}. CLIP’s text encoder is used to enable multimodal joint conditioning and for compatibility with Stable Diffusion's embedding space. Using ViT-L/14 \cite{dosovitskiy2021image}, the hidden state contains $257$ channels for the image encoder and $77$ channels for the text encoder. To facilitate identity preservation, we use the last hidden state of the image and text encoders instead of the final encoded vector, which maintains an associated channel dimension \cite{kulal2023putting}. A linear layer projects the image encoder's last hidden state from 1024 dimensions to 768 dimensions, to match the text encoder.

To enable pose conditioning we use 2D, 17-DOF skeletons. To obtain the poses, RTMPose was selected over OpenPose as it performs well on the blurry and complex scenes found in Kinetics. Each pose keypoint contains a predicted $x,y$ coordinate as well as a confidence score. We concatenate and flatten the pose into a $(1,51)$ vector and project it through a linear layer into a $768$ dimension embedding space. The pose, image, and text embeddings are concatenated into a $(batch 
size, 335, 768)$ tensor.

At training time, we finetune the Stable Diffusion checkpoint with four configurations:

\begin{itemize}
\item C1 ``baseline'' Finetuned on only the encoded reference images as conditioning.

\item C2 ``pose-only'' Finetuned on the encoded reference images with pose as conditioning.

\item C3 ``text-only'' Finetuned on the encoded reference images with text as conditioning.

\item C4 ``pose+text'' Finetuned on the combined encoded reference images, text, and pose as conditioning.
\end{itemize}

Each configuration was trained for approximately $600$ epochs.

\subsection{Evaluation}
Our objective is to enable complex non-rigid edits driven by text prompts on in-the-wild data. In addition, we are interested in improving the quality and contextual awareness of person-object interactions.

To quantify the realism of the generated images, we first evaluate on traditional metrics including FID (photorealism) \cite{heusel2018fid} and PCKh (pose adherence) \cite{andriluka2014pose} across all four model configurations. The numerical results are shown in Table \ref{tab:FID,PCKh}. However, the metrics are imperfect for measuring pose preservation across two images in a scene. While FID measures photorealism, the metric does not enforce identity preservation; and while PCKh measures pose adherence, our goal is to hallucinate plausible but controllable poses. Therefore, a model can learn an effective distribution for how to insert a person, but receive unremarkable PCKh scores. In absence of a metric to quantify identity preservation and pose believability, we conduct a user study to obtain human ratings for each configuration. The user study protocol was submitted for approval  by <anonymized>'s Institute Review Board.

\begin{table}[t]
\centering
\begin{minipage}{0.45\textwidth}
\centering
\begin{tabular}{|c|c|c|}
\hline
Config & Identity & Controllable \\ \hline
Img-Only (Kulal* \cite{kulal2023putting})      & 61\% & N/A \\ \hline
Img, Text     & 55\% & 39\% \\ \hline
Img, Pose     & 63.5\% & \textbf{57.5\%} \\ \hline
Img, Pose, Text & \textbf{68.5\%} & 51\% \\ \hline
\end{tabular}
\caption{User study on photos \textbf{without object interactions}. Adding text or pose to image conditioning improves identity preservation and helps respect the control signal. We reimpliment Kulal et al. on publicly available data, as denoted by "*". This provides a fair comparison as much of their training data is unavailable. As you can see from Figure \ref{fig:nonrigid_comparison}, the photorealism goes down. We trade a small amount of photorealism for controllability in generations. }
\label{tab:user_study_non_object}
\end{minipage}%
\hspace{0.05\textwidth} 
\begin{minipage}{0.45\textwidth}
\centering
\begin{tabular}{|c|c|c|}
\hline
Config & Identity & Interactions \\ \hline
Img-Only (Kulal*) \cite{kulal2023putting}      & \textbf{50\%} & 24\% \\ \hline
Img-Text      & 29.5\% & 24.5\% \\ \hline
Img-Pose      & 25\% & \textbf{33\%} \\ \hline
Img-Pose-Text & 41\% & \textbf{33\%} \\ \hline
\end{tabular}
\caption{User study on photos \textbf{with person-object interactions}. Identity preservation is harder on person-object scenes. Adding pose and or text creates better object interactions. We reimpliment Kulal et al. on publicly available data, as denoted by "*". This provides a fair comparison as much of their training data is unavailable.}
\label{tab:user_study_object}
\end{minipage}
\end{table}

Eight human raters were recruited from the university's machine learning center. Raters were presented scenes with the original image, the text caption, the pose skeleton, the generated image, and a visualization of the ground truth skeleton overlayed onto the generated image.  Raters were asked two questions per scene, and responded to each question with score of either 0 or 1, denoting whether or not the identity was preserved, whether or not the controlling signal was adhered to, and whether or not object-person interactions were plausible.

For all scenes, raters were asked how well the generated image preserves the appearance of the subject. For scenes with person-object interactions, raters were asked how reasonable the generated image's interaction is between the person and object. For scenes without object interactions, raters were asked how well the generated image respects the input caption or pose. In total, $50$ scenes were rated from each of the four configurations.

\begin{figure}[t!]
\centering
\begin{tabular}{@{\hskip 0.5pt}c@{\hskip 0.5pt}@{\hskip 0.5pt}c@{\hskip 0.5pt}@{\hskip 10pt}c@{\hskip 0.5pt}@{\hskip 10pt}c@{\hskip 0.5pt}@{\hskip 0.5pt}c@{\hskip 0.5pt}}
Scene & Ref & GT Target & Img-Pose & Img-Pose-Text \\

\includegraphics[width=0.17\columnwidth]{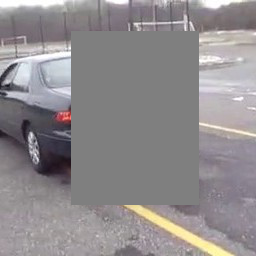} &
\includegraphics[width=0.17\columnwidth]{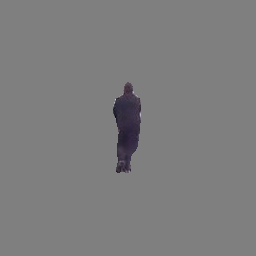} &
\includegraphics[width=0.17\columnwidth]{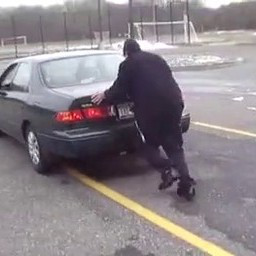} &
\includegraphics[width=0.17\columnwidth]{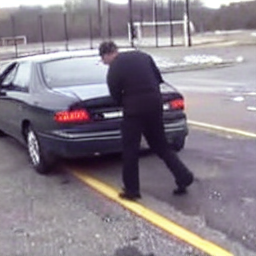} &
\includegraphics[width=0.17\columnwidth]{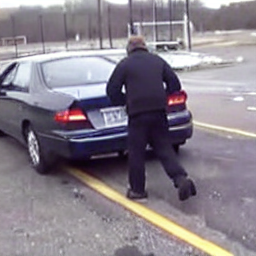} \\
\multicolumn{5}{p{0.95\columnwidth}}{\footnotesize\textit{Row 1: "He approaches the car and places his hands on the trunk."}} \\

\includegraphics[width=0.17\columnwidth]{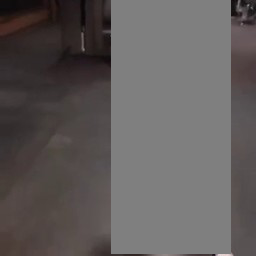} &
\includegraphics[width=0.17\columnwidth]{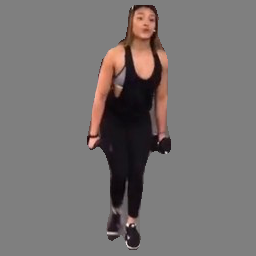} &
\includegraphics[width=0.17\columnwidth]{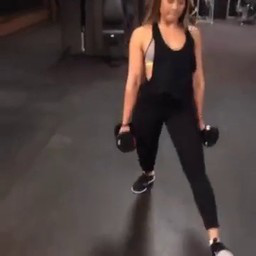} &
\includegraphics[width=0.17\columnwidth]{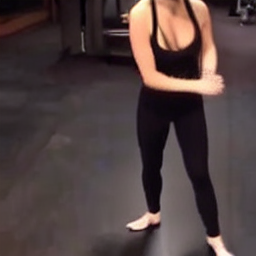} &
\includegraphics[width=0.17\columnwidth]{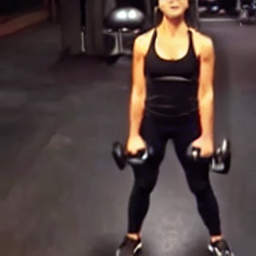} \\
\multicolumn{5}{p{0.95\columnwidth}}{\footnotesize\textit{Row 2: "She steps out to her right side while holding dumbbells."}} \\

\includegraphics[width=0.17\columnwidth]{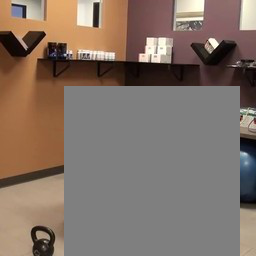} &
\includegraphics[width=0.17\columnwidth]{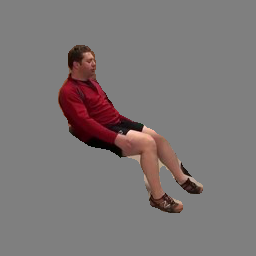} &
\includegraphics[width=0.17\columnwidth]{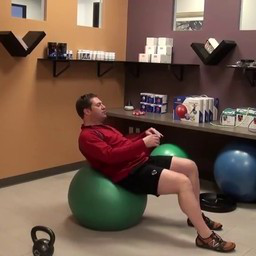} &
\includegraphics[width=0.17\columnwidth]{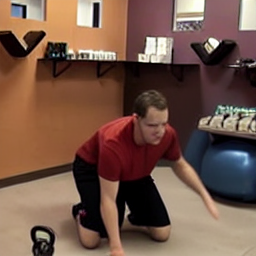} &
\includegraphics[width=0.17\columnwidth]{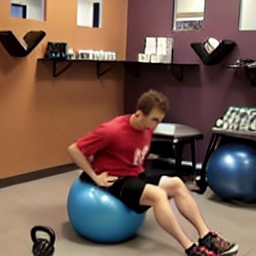} \\
\multicolumn{5}{p{0.95\columnwidth}}{\footnotesize\textit{Row 3: "He leans back slightly, engaging his core muscles while sitting on the exercise ball."}} \\

\includegraphics[width=0.17\columnwidth]{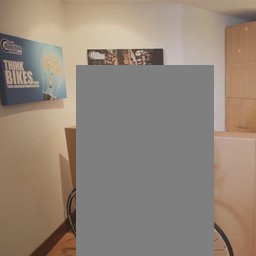} &
\includegraphics[width=0.17\columnwidth]{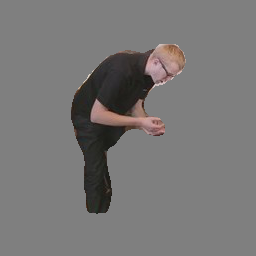} &
\includegraphics[width=0.17\columnwidth]{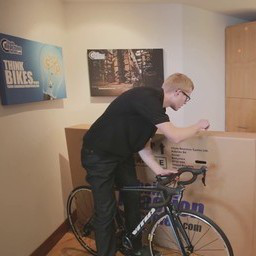} &
\includegraphics[width=0.17\columnwidth]{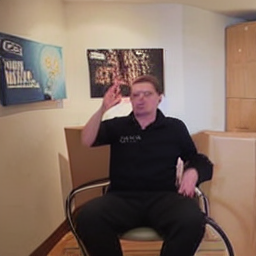} &
\includegraphics[width=0.17\columnwidth]{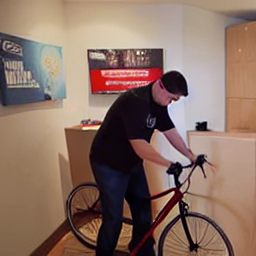} \\
\multicolumn{5}{p{0.95\columnwidth}}{\footnotesize\textit{Row 4: "He leans forward, placing his hands further down onto the bike’s handlebars."}} \\

\includegraphics[width=0.17\columnwidth]{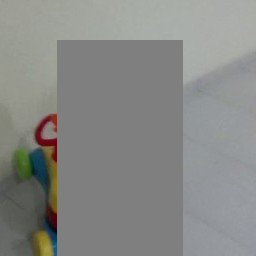} &
\includegraphics[width=0.17\columnwidth]{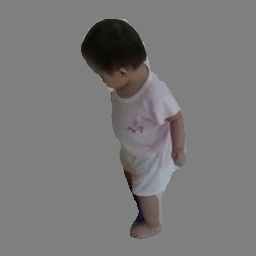} &
\includegraphics[width=0.17\columnwidth]{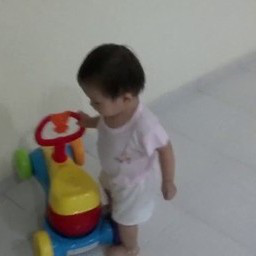} &
\includegraphics[width=0.17\columnwidth]{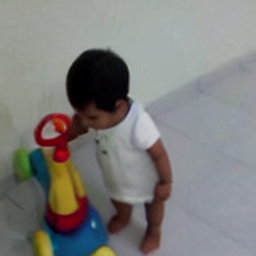} &
\includegraphics[width=0.17\columnwidth]{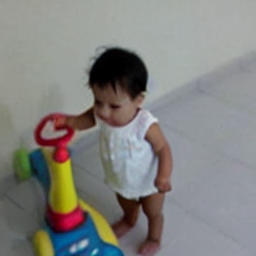} \\
\multicolumn{5}{p{0.95\columnwidth}}{\footnotesize\textit{Row 5: "She turns slightly to her right, looking down at the toy."}} \\

\end{tabular}
\caption{Using pose conditioning is insufficient to specify person-object interactions. Combining textual embeddings allows the model to contextualize how the person interacts with their surroundings. For example, the exercise ball or bicycle are faithfully maintained with text on Rows 3 and 4, but not with just pose. We show qualitative improvement of person-object interactions by conditioning on weakly generated captions, combined with reference images and pose data. We use the pose from the ground truth combined with the pose from the reference image for conditioning, fed through a single learnable dense layer.}
\label{fig:image_text_pose_vs_image_pose}
\end{figure}

We study how image, text, and pose combined qualitatively improve object interactions in Figure \ref{fig:image_text_pose_vs_image_pose}.

\subsection{Data Processing}

In order to finetune the inpainting diffusion network, we first have to process video data into a series of self-supervised frame pairs. We found that particularly in the context of joint conditioning, the sampling criteria and procedure are very important for performance. We study Charades, Fit3D, NTU-RGBD60, and Kinetics. We apply our filtering procedure described in methods and retain between a fraction of videos depending on the dataset. We condense clips down to 2-5 frames, depending on our minimum edit distance criteria. We also experimented with retaining more clips from NTU and found that the model learned poor action distributions and would mostly guess the same identity pose, as most of the time in videos, people are just standing or sitting without performing actions. We follow the same masking and augmentation procedures as \cite{kulal2023putting}.

\subsection{Controllable Non-Rigid Editing in the Wild}

We first study our objective of enabling controllable, non-rigid image edits in the wild. Previous work has focused on letting the model decide the best way to insert a person into a scene without human-in-the-loop control. While \cite{cao2023masactrl} enables image-text control, it does so without finetuning and is brittle on real-world data, as we show in Figure \ref{fig:nonrigid_comparison}. Similarly, \cite{bhunia2023person} allows img-pose control, but is also brittle on real-world data. Figure \ref{fig:multiple_edits_same_photo} and Figure \ref{fig:complex_nonrigid} show examples.

We study traditional diffusion quantitative metrics in Table \ref{tab:FID,PCKh}. Our image, pose and image-text, pose conditioned models perform the best in terms of PCKh, which measures adherence to the ground truth pose. The image-only model performs best on FID, which measures photo-realism. However, both traditional diffusion quantitative metrics may not be well-suited to fairly evaluate performance on our task. 

\subsection{Person-Object Interactions}
In addition, Table \ref{tab:user_study_non_object} studies the faithfulness to control signals for image-pose, image-text, and image-pose-text models. That is, given a scene change caption, users are asked to rate each image on whether the generation follows the caption. Similarly, given a target pose visualization, users are asked to evaluate if it is a plausible pose generation. Lastly, in Table \ref{tab:user_study_object} we also study the quality of Object-Person interactions. Users are asked to make a binary decision for each scene, on whether the interaction with each object is plausible. Responses are averaged for 4 users each into the tables below.

\subsection{Comparisons to Baselines}

In Figure \ref{fig:nonrigid_comparison}, we compare our method to several existing baseline methods.
PIDM is a leading pose-conditioned image editing model but contains modules handcrafted for the simple datasets it was trained on, failing on textures from ITW pose-editing edits. The demo for PIDM is limited to a set of predefined poses, so we include results in the appendix to have a fair comparison.

For the comparison to MASACtrl, they accept only a single image at inference time, so we provide the reference image. Our model accepts the reference image to preserve the identity of, as well as the target image masked with a bounding box to inpaint.

\section*{Discussion}

\subsection{Limitations}

Despite the promising results, our approach has several limitations. The model often fails to preserve the identity of the person being edited, as observed in user studies where identity preservation rates range from 55\% to 68.5\%, depending on the conditioning configuration. The image-text and image-pose models achieve the best results due to additional relevant conditioning information. The model is brittle on actions that involve long, narrow objects. For scenes with objects, identity preservation occurring only 25-50\% of the time. Additionally, object-person compatible hallucinations occur only 24-33\% of the time, with the best results achieved by the image-pose and image-pose-text models. See Failure Cases for qualitative demonstrations. We hypothesize that results would improve for the image-text model with more accurate and descriptive captions, as scene difference captions represent a different task than pretraining. The estimated cost of training our model is approximately \$300, given the use of 8 Nvidia A40 GPUs for 48 hours at current cloud prices starting at \$0.79 per hour.

\subsection{Broader Impact}
This research has the potential to advance the field of image editing by providing more intuitive and user-friendly tools for complex, non-rigid edits. This approach could also lead to better video generations and be used to create data to study human affordances more broadly, or to evaluate affordance knowledge learned by diffusion models. However, it also raises ethical concerns regarding the misuse of such technology for creating deepfakes or other misleading content. Additionally, the dependence on large datasets and powerful computational resources could exacerbate existing inequalities in AI research.  As such, it is crucial to implement safeguards and promote responsible use to mitigate potential negative impacts.

\section*{Conclusion}
In this paper, we presented a novel method for controllable and complex, non-rigid image edits using multimodal conditioning and self-supervised learning from video datasets on in the wild data. Our approach demonstrates significant improvements in identity preservation and contextual accuracy in generated images, particularly in scenes involving human-object interactions. While our method shows promise, further research is needed to address its limitations and ensure ethical and responsible deployment. Overall, our work contributes to the advancement of intuitive, user-centric image editing technologies and sets the stage for future developments in this field.

\bibliographystyle{plain}
\bibliography{references}

\clearpage 

\appendix

\end{document}


\appendix

\section{Appendices}

\subsection{Information About Our Data}

We describe statistics on the datasets we train on in Table \ref{tab:dataset_summary}. 

\begin{table*}[t]
\centering
\begin{tabular}{|p{3cm}|c|c|c|c|c|}
\hline
Dataset & Orig Videos & Videos & Frames & Captions & Caption Length \\ \hline
Kinetics \cite{carreira2019kinetics} & 650,000 & 13,400 & 42,934 & 5,787 & 68.8 \\ \hline
NTU-RGB+D \cite{Shahroudy_2016_NTU_RGBD} & 57,600 & 65,775 & 3,412,398 & 7,700 & 91.7 \\ \hline
\end{tabular}
\caption{Summary statistics for datasets we annotated with captions and trained on. We denote the original number of videos versus the number of videos "Videos" after our preprocessing. Note that we followed the original pipeline of Kulal et al for NTU, to test how identity preservation responds to this condition. Note that the number of videos went up after filtering for NTU, because the default pipeline splits videos into "sub-videos," and has much looser guidelines for filtering.}
\label{tab:dataset_summary}
\end{table*}

\begin{figure}[t!]
\centering
\begin{tabular}{@{\hskip 0.5pt}c@{\hskip 0.5pt} @{\hskip 0.5pt}c@{\hskip 0.5pt} @{\hskip 0.5pt}c@{\hskip 0.5pt} @{\hskip 0.5pt}c@{\hskip 0.5pt}}
Scene & Ref & Target & Pose Det \\
\includegraphics[width=0.23\columnwidth]{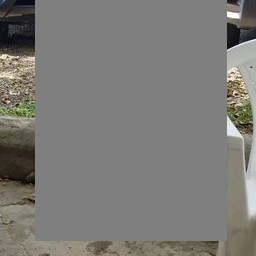} &
\includegraphics[width=0.23\columnwidth]{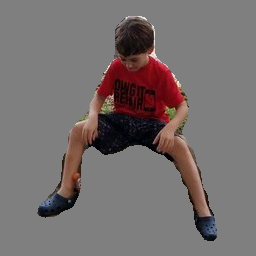} &
\includegraphics[width=0.23\columnwidth]{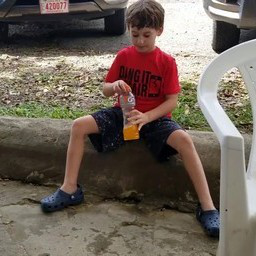} &
\includegraphics[width=0.23\columnwidth]{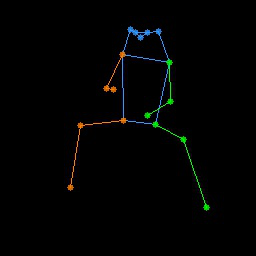}  \\
\includegraphics[width=0.23\columnwidth]{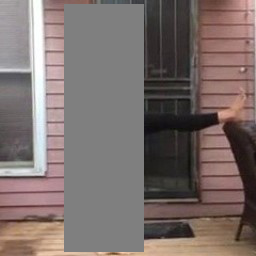} &
\includegraphics[width=0.23\columnwidth]{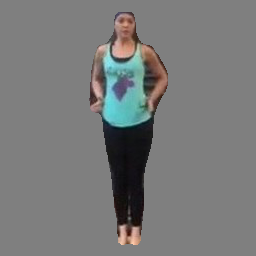} &
\includegraphics[width=0.23\columnwidth]{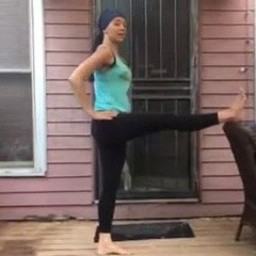} &
\includegraphics[width=0.23\columnwidth]{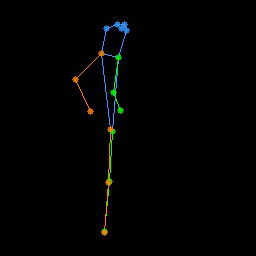}  \\
\includegraphics[width=0.23\columnwidth]{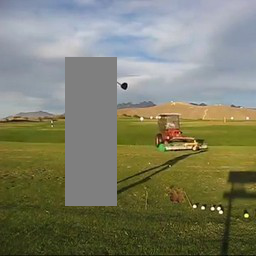} &
\includegraphics[width=0.23\columnwidth]{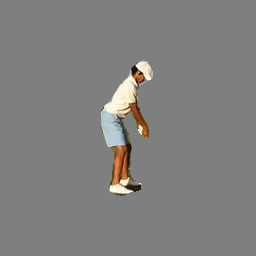} &
\includegraphics[width=0.23\columnwidth]{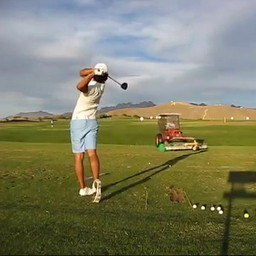} &
\includegraphics[width=0.23\columnwidth]{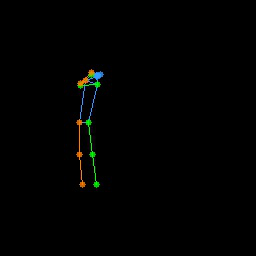} \\
\end{tabular}
\caption{Our data encompasses a variety of action classes and scenes with and without object interactions from Kinetics. We sample a few key frames per video, according to the minimum pose distance criteria described in Methods, which allows us to learn meaningful policies from much more limited data than comparable methods.\\
Row 1: \textit{"He raises the bottle slightly, looking at it."} \\
Row 2: \textit{"She raises her left leg to the side while placing her left hand on her hip."} \\
Row 3: \textit{"He follows through with his swing, raising the golf club over his shoulder."} }
\label{fig:DataExamples}
\end{figure}

We showcase examples of our data in Figure \ref{fig:DataExamples}

\subsection{Failure Cases}
\begin{figure}[h!]
\centering
\begin{tabular}{@{\hskip 0.5pt}c@{\hskip 0.5pt} @{\hskip 0.5pt}c@{\hskip 0.5pt} @{\hskip 0.5pt}c@{\hskip 0.5pt} @{\hskip 0.5pt}c@{\hskip 0.5pt} @{\hskip 0.5pt}c@{\hskip 0.5pt}}
 \\
\includegraphics[width=0.16\columnwidth]{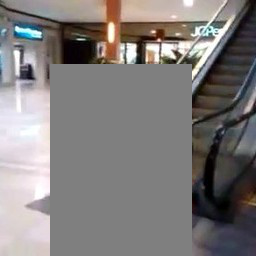} & 
\includegraphics[width=0.16\columnwidth]{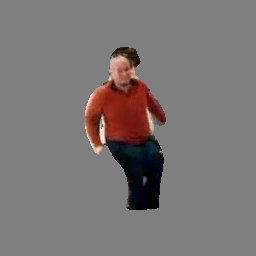} & 
\includegraphics[width=0.16\columnwidth]{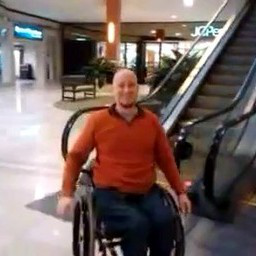} & 
\includegraphics[width=0.16\columnwidth]{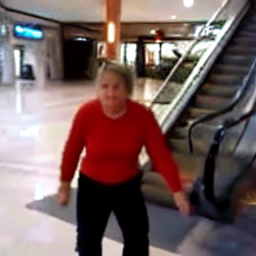} & 
\includegraphics[width=0.16\columnwidth]{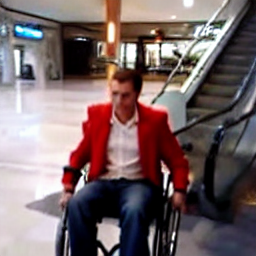} \\
\includegraphics[width=0.16\columnwidth]{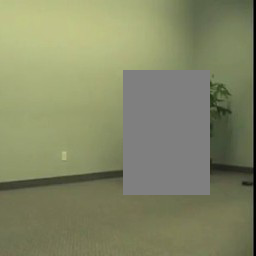} & 
\includegraphics[width=0.16\columnwidth]{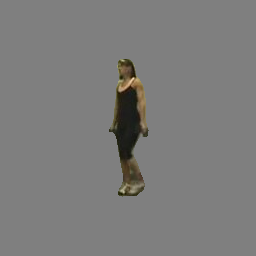} & 
\includegraphics[width=0.16\columnwidth]{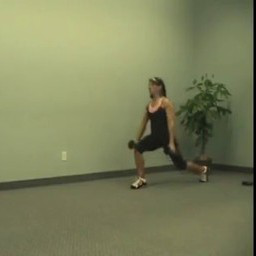} & 
\includegraphics[width=0.16\columnwidth]{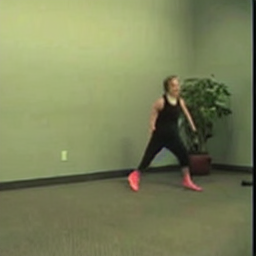} & 
\includegraphics[width=0.16\columnwidth]{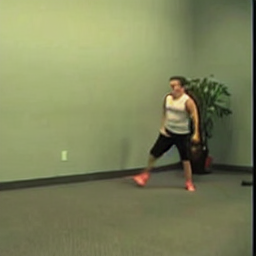} \\
\includegraphics[width=0.16\columnwidth]{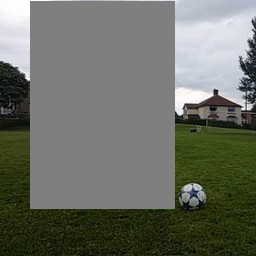} & 
\includegraphics[width=0.16\columnwidth]{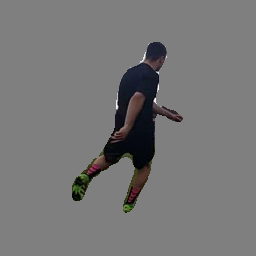} & 
\includegraphics[width=0.16\columnwidth]{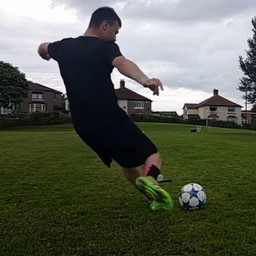} & 
\includegraphics[width=0.16\columnwidth]{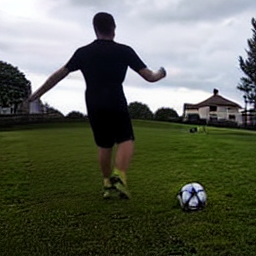} & 
\includegraphics[width=0.16\columnwidth]{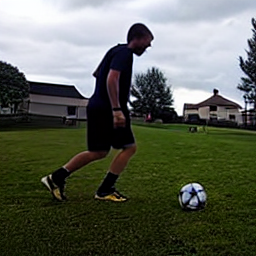} \\
\includegraphics[width=0.16\columnwidth]{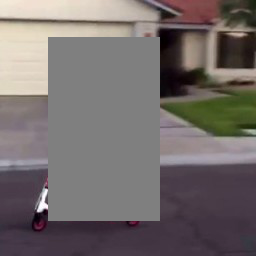} & 
\includegraphics[width=0.16\columnwidth]{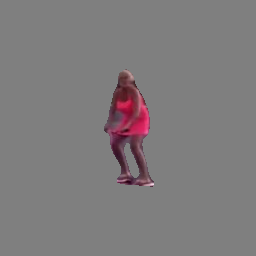} & 
\includegraphics[width=0.16\columnwidth]{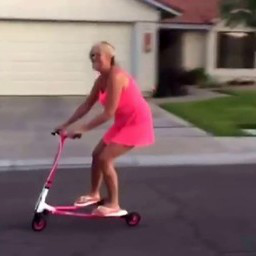} & 
\includegraphics[width=0.16\columnwidth]{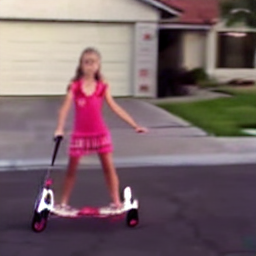} & 
\includegraphics[width=0.16\columnwidth]{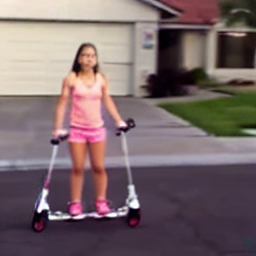} \\
\includegraphics[width=0.16\columnwidth]{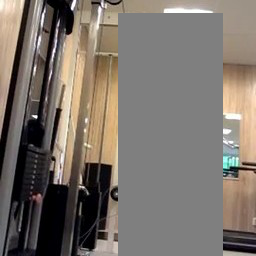} & 
\includegraphics[width=0.16\columnwidth]{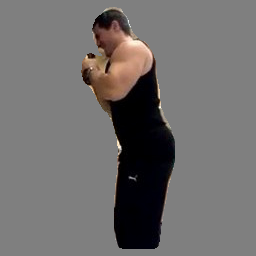} & 
\includegraphics[width=0.16\columnwidth]{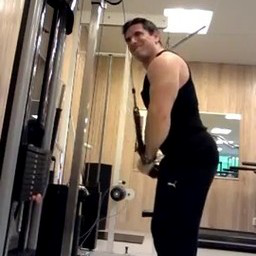} & 
\includegraphics[width=0.16\columnwidth]{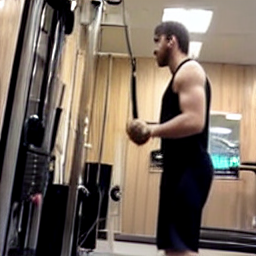} & 
\includegraphics[width=0.16\columnwidth]{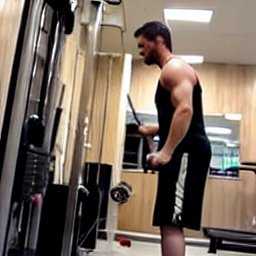} \\
\end{tabular}
\caption{
Columns: Masked target image, Reference image , Ground Truth, Img-Pose, Img-Pose-Text
Failure cases of our model to preserve identity, object interaction, or understand semantics of the scene. The img-pose model can sometimes fail to capture the context of the scene, such as kicking a soccer ball, or doing an exercise lunge, or generation with a wheelchair but text, img, pose can help to rectify it. Sometimes, both models fail to capture a good object interaction, particularly if it is mostly obfuscated or not common in the dataset. The model often fails with interactions with long slender objects, such as the cable machine or an axe.}
\label{fig:Failure_Cases}
\end{figure}

We include a few results of failure cases for our model in \ref{fig:Failure_Cases}.

\subsection{Efficacy of Captioning Pipeline}
We study the efficacy of our captioning pipeline in \ref{fig:caption_examples}
\label{fig:caption_examples}
\begin{figure}[t!]
\centering
\begin{tabular}{@{\hskip 0.5pt}c@{\hskip 0.5pt} @{\hskip 0.5pt}c@{\hskip 0.5pt}}
\includegraphics[width=0.45\columnwidth]{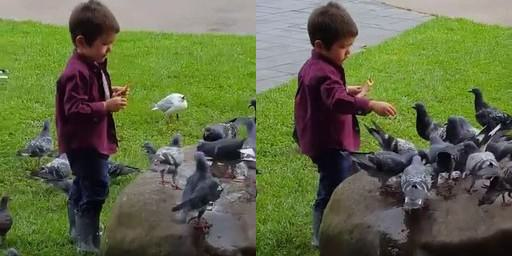} &
\includegraphics[width=0.45\columnwidth]{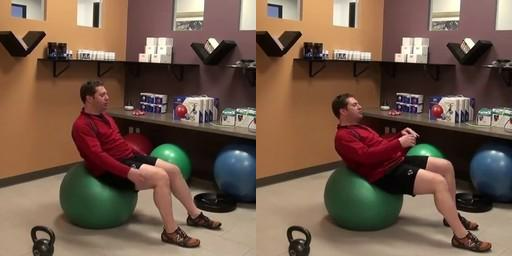} \\
\includegraphics[width=0.45\columnwidth]{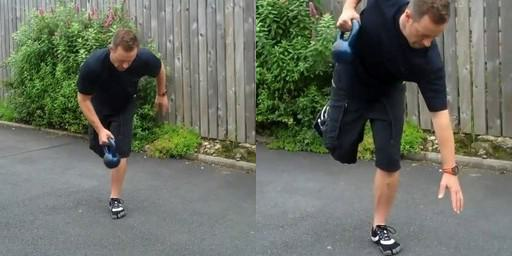} &
\includegraphics[width=0.45\columnwidth]{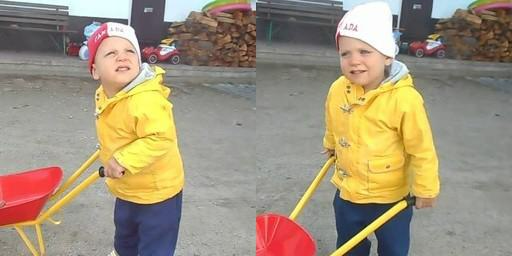} \\
\end{tabular}
\caption{Our supervision is noisy, yet we achieve good signs of life for controllability. Examples of generated captions: (Top Left) "He bends over slightly and extends his arm toward the birds." (Top Right) "He leans back slightly engaging his core muscles while sitting on the exercise ball." (Middle Left) "He lifts his right foot off the ground, leaning forward." (Middle Right) "He shifts his gaze from looking up to looking straight ahead." Additional data captioning could create robust models. Most generations capture the essence of the change of motion. However, caption generations are limited by sampling (despite filtering, sometimes there is a small change in pose). Also, sometimes the model captions the two images in reverse order or misidentifies uncommon objects like an indoor trampoline, and it often reverses left and right.}
\end{figure}

\subsection{Examples of qualitative adherence to pose}
\begin{figure}[t!]
\centering
\begin{tabular}{@{\hskip 0.5pt}c@{\hskip 0.5pt} @{\hskip 0.5pt}c@{\hskip 0.5pt} @{\hskip 0.5pt}c@{\hskip 0.5pt} @{\hskip 0.5pt}c@{\hskip 0.5pt}}
Masked Target & Reference & Ground Truth & Pose Overlay \\
\includegraphics[width=0.22\columnwidth]{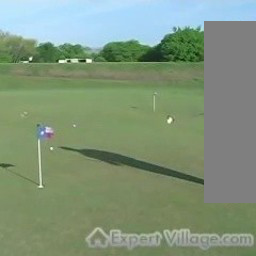} & 
\includegraphics[width=0.22\columnwidth]{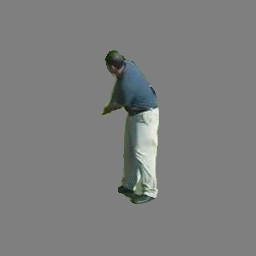} & 
\includegraphics[width=0.22\columnwidth]{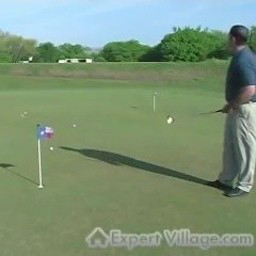} & 
\includegraphics[width=0.22\columnwidth]{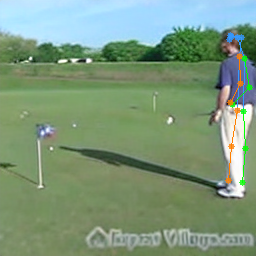} \\
\includegraphics[width=0.22\columnwidth]{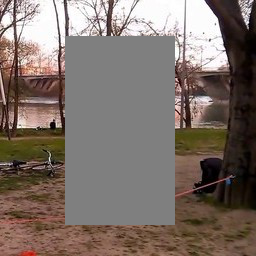} & 
\includegraphics[width=0.22\columnwidth]{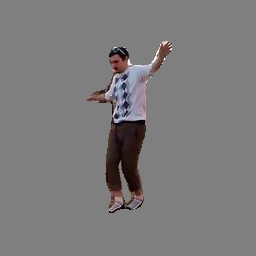} & 
\includegraphics[width=0.22\columnwidth]{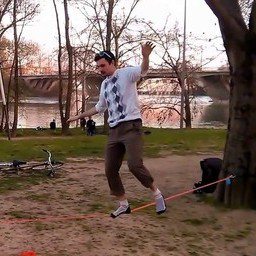} & 
\includegraphics[width=0.22\columnwidth]{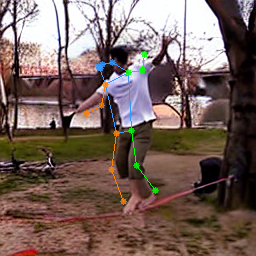} \\
\includegraphics[width=0.22\columnwidth]{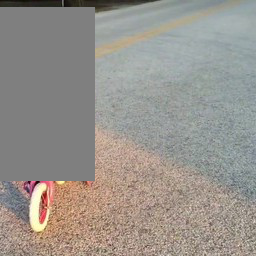} & 
\includegraphics[width=0.22\columnwidth]{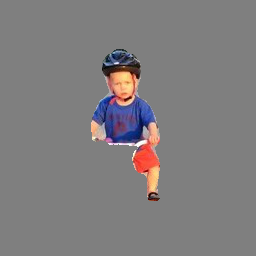} & 
\includegraphics[width=0.22\columnwidth]{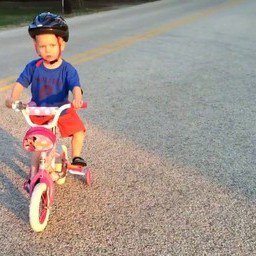} & 
\includegraphics[width=0.22\columnwidth]{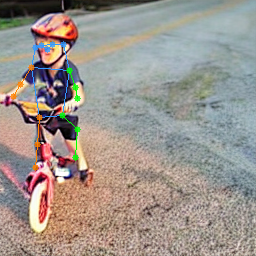} \\
\includegraphics[width=0.22\columnwidth]{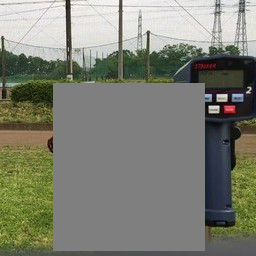} & 
\includegraphics[width=0.22\columnwidth]{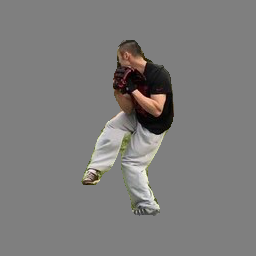} & 
\includegraphics[width=0.22\columnwidth]{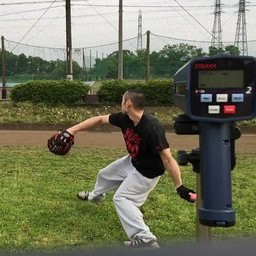} & 
\includegraphics[width=0.22\columnwidth]{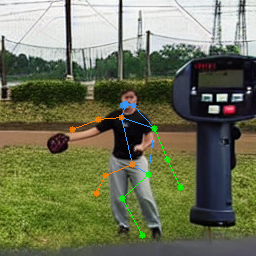} \\
\end{tabular}
\caption{Pose conditioning works well in various scenarios. Visualization of pose adherence for the image-pose model. The model generally adheres well to pose conditioning but can struggle with complex scenes, backgrounds, and many object-person interactions. Larger filtered datasets could help alleviate some of these issues.}
\label{fig:pose_adherence}
\end{figure}

We study the efficacy of pose-conditioning in Figure \ref{fig:pose_adherence}.

\subsection{PIDM Failure Cases}
\begin{figure}[t!]
\centering
\begin{tabular}{@{\hskip 0.5pt}c@{\hskip 0.5pt}@{\hskip 0.5pt}c@{\hskip 0.5pt}@{\hskip 0.5pt}c@{\hskip 0.5pt}@{\hskip 0.5pt}c@{\hskip 0.5pt}@{\hskip 0.5pt}c@{\hskip 0.5pt}}
Input Image & Pose 1 & Pose 2 & Pose 3 & Pose 4 \\
\includegraphics[width=0.18\columnwidth]{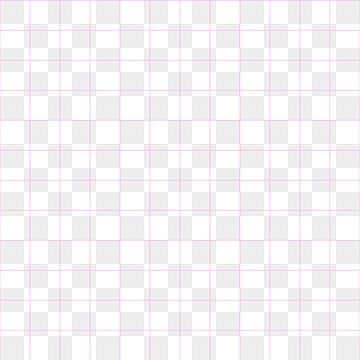} &
\includegraphics[width=0.18\columnwidth]{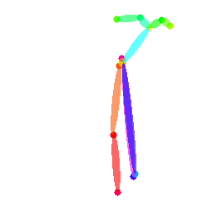} &
\includegraphics[width=0.18\columnwidth]{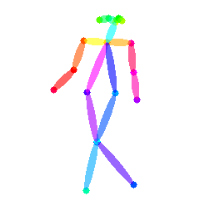} &
\includegraphics[width=0.18\columnwidth]{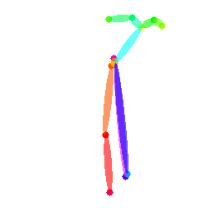} &
\includegraphics[width=0.18\columnwidth]{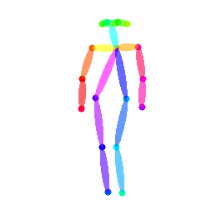} \\

\includegraphics[width=0.18\columnwidth]{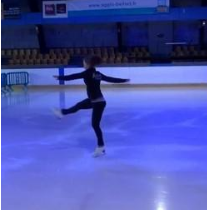} &
\includegraphics[width=0.18\columnwidth]{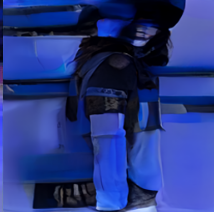} &
\includegraphics[width=0.18\columnwidth]{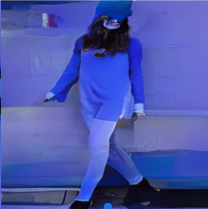} &
\includegraphics[width=0.18\columnwidth]{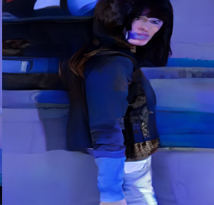} &
\includegraphics[width=0.18\columnwidth]{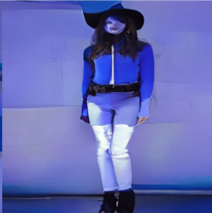} \\

\includegraphics[width=0.18\columnwidth]{Figure8_PIDM/gridded_blank.jpg} &
\includegraphics[width=0.18\columnwidth]{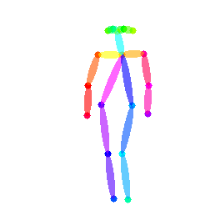} &
\includegraphics[width=0.18\columnwidth]{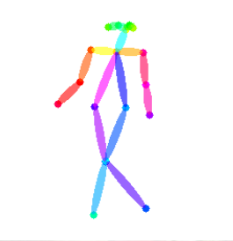} &
\includegraphics[width=0.18\columnwidth]{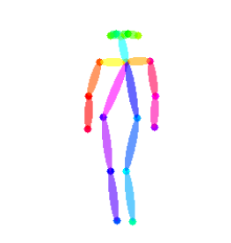} &
\includegraphics[width=0.18\columnwidth]{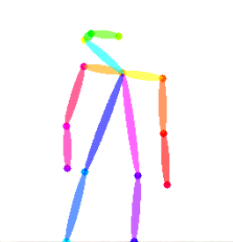} \\

\includegraphics[width=0.18\columnwidth]{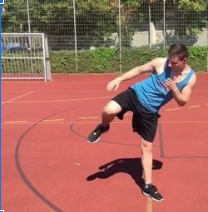} &
\includegraphics[width=0.18\columnwidth]{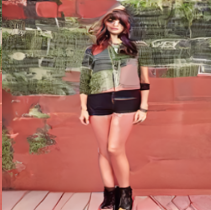} &
\includegraphics[width=0.18\columnwidth]{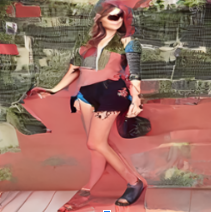} &
\includegraphics[width=0.18\columnwidth]{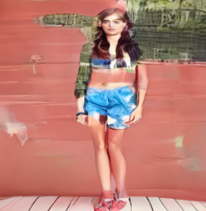} &
\includegraphics[width=0.18\columnwidth]{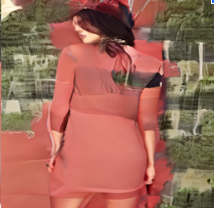} \\
\end{tabular}
\caption{Failure cases of the PIDM \cite{bhunia2023person} model on "In-The-Wild" data across various poses and input images. Rows 1 and 3 represent different pose variations. Our model, in contrast, is trained to maintain identity preservation and generate plausible, controllable results in more complex scenes.}
\label{fig:PIDM}
\end{figure}

We study examples of failures in PIDM \cite{bhunia2023person}, a SOTA pose-conditioned image editing model in Figure \ref{fig:PIDM}.

\clearpage

\bibliographystyle{plain}
\bibliography{references}